\documentclass{article}
\usepackage{amssymb,latexsym,graphicx, amsmath} 

\usepackage{etoolbox}
\let\bbordermatrix\bordermatrix
\patchcmd{\bbordermatrix}{8.75}{4.75}{}{}
\patchcmd{\bbordermatrix}{\left(}{\left[}{}{}
\patchcmd{\bbordermatrix}{\right)}{\right]}{}{}

\usepackage{algorithmic}
\usepackage{algorithm}
\usepackage{caption}
\usepackage{subcaption}
\usepackage{float}
\usepackage{placeins}

\begin{document}

\begin{center}
\huge A Case Study in Text Mining: Interpreting Twitter Data From World Cup Tweets 

\end{center}

\vspace{0.1in}

\begin{center}
Daniel Godfrey \footnote{Department of Mathematics, University of North Carolina, Charlotte, NC 28223, USA 

(dgodfre4@uncc.edu)}, Caley Johns \footnote{Department of Mathematics, Brigham Young University -- Idaho, ID 83440, USA 

(joh11066@byui.edu)}, 
Carol Sadek \footnote{Department of Mathematics, Wofford College, Spartanburg, SC 29303, USA 

(sadekcw@email.wofford.edu)},

Carl Meyer \footnote{Department of Mathematics, NC State University, Raleigh, NC 27695, USA

(meyer@ncsu.edu)},
 Shaina Race \footnote{Department of Mathematics, NC State University, Raleigh, NC 27695, USA
 
 (slrace@ncsu.edu)}

\end{center}

\vspace{0.1in}

\begin{abstract}

Cluster analysis is a field of data analysis that extracts underlying patterns in data.
One application of cluster analysis is in text-mining, the analysis of large collections of text to find similarities between documents. 
We used a collection of about 30,000 tweets extracted from Twitter just before the World Cup started. 
A common problem with real world text data is the presence of  linguistic noise. 
In our case it would be extraneous tweets that are unrelated to dominant themes.
To combat this problem, we created an algorithm that combined the DBSCAN algorithm and a consensus matrix.
This way we are left with the tweets that are  related to those dominant themes. 
We then used cluster analysis to find those topics that the tweets describe.
We clustered the tweets using $k$-means, a commonly used clustering algorithm, and Non-Negative Matrix Factorization (NMF) 
and compared the results. 
The two algorithms gave similar results, but NMF proved to be faster and provided more easily interpreted results.
We explored our results using two visualization tools, Gephi and Wordle. 

\end{abstract}



\let \thefootnote \relax \footnote{* This research was supported in part by NSF Grant DMS-1063010 and NSA Grant H98230-12-1-0299}




\section{Background Information}

Cluster analysis is the process of grouping data points together based on their relative similarities. 
Text mining, a subfield of cluster analysis, is the analysis of large collections of text to find patterns between documents. 

\vspace{0.2in}

We extracted tweets from Twitter containing the words `world cup'; this was before the World Cup games had started.
In the beginning we had 29,353 tweets.
The tweets consisted of English and Spanish words.
After working with the data we kept 17,023 tweets that still contained the important information.

\vspace{0.2in}

Twitter is a useful tool for gathering information about its users' demographics and their opinions about certain subjects. 
For example, a political scientist could see what a younger audience feels about certain news stories, or an advertiser could find out what Twitter users are saying about their products. 
With security it is important to be able to discern between threats and non threats.
Search engines also use this to discern between the various topics that can apply to one word.
For example, `Jordan' could apply to Michael Jordan, the country Jordan, or the Jordan River.

\vspace{0.2in}

There are many different clustering algorithms each with their advantages and disadvantages. 
No clustering algorithm is perfect and each provides a slightly different clustering.
Certain clustering algorithms are better with certain types of datasets such as text data or numerical data or data with a wide range of cluster sizes. 
The purpose of this research is to compare advantages and disadvantages of two clustering algorithms, Non-Negative Matrix Factorization and the more widely used $k$-means algorithm, when used on Twitter text data.

\section{Algorithms}

\subsection{$k$-Means}

One way to cluster data points is through an algorithm called $k$-means, the most widely used algorithm in the field. 
The purpose of this algorithm is to divide $n$ data points into $k$ clusters where the distance between each data point and its cluster's center is minimized. 
Initially $k$-means chooses $k$ random points from the data space, not necessarily points in the data, and assigns them as centroids. 
Then, each data point is assigned to the closest centroid to create $k$ clusters. 
After this first step, the centroids are reassigned to minimize the distance between them and all the points in their cluster.
Each data point is reassigned to the closest centroid. 
This process continues until convergence is reached. 

\vspace{0.2in}

The distance between data points and centroids can be measured using several different metrics including the most widely used cosine and Euclidean distances \cite{Dist}. Cosine distance is a measure of distance between two data points while Euclidean distance is the magnitude of the distance between the data points. For example, if two data points represented two sentences both containing three words in common, cosine would give them a distance that is independent from how many times the three words appear in each of the sentences. Euclidean distance, however takes into account the magnitude of similarity between the two sentences. Thus, a sentence containing the words ``world'' and ``cup'' 3 times and another containing those words 300 times are considered more dissimilar by Euclidean distance than by cosine distance.

\vspace{0.2in}

For the purpose of this research, we used cosine distance because it is faster, better equipped to dealing with sparse matrices, and provides distances between tweets that are independent of the tweets' lengths. Thus, a long tweet with several words might still be considered very similar to a shorter tweet with fewer words. Cosine distance measures the cosine of the angle between two vectors such that
\[
\cos \theta = \frac{x \cdot y}{\lVert x \rVert \lVert y \rVert},
\]
where $x$ and $y$ are term frequency-inverse document frequency (TF-IDF) vectors corresponding to documents $x$ and $y$. The resulting distance ranges from $-1$ to 1. However, since $x$ and $y$ are vectors that contain all non-negative values, the cosine distance ranges from 0 to 1.

\vspace{0.2in}

One of the disadvantages of $k$-means is that it is highly dependent on the initializations of the centroids. 
Since these initializations are random, multiple runs of $k$-means produce different results \cite{kMeansAnim}. 
Another disadvantage is that the value of $k$ must be known in order to run the algorithm.
 With real-world data, it is sometimes difficult to know how many clusters are needed before performing the algorithm.

\subsection{Consensus Clustering}

\begin{figure}[h]
        \centering
          \begin{subfigure}[b]{0.48\textwidth}
	\includegraphics[width=\textwidth]{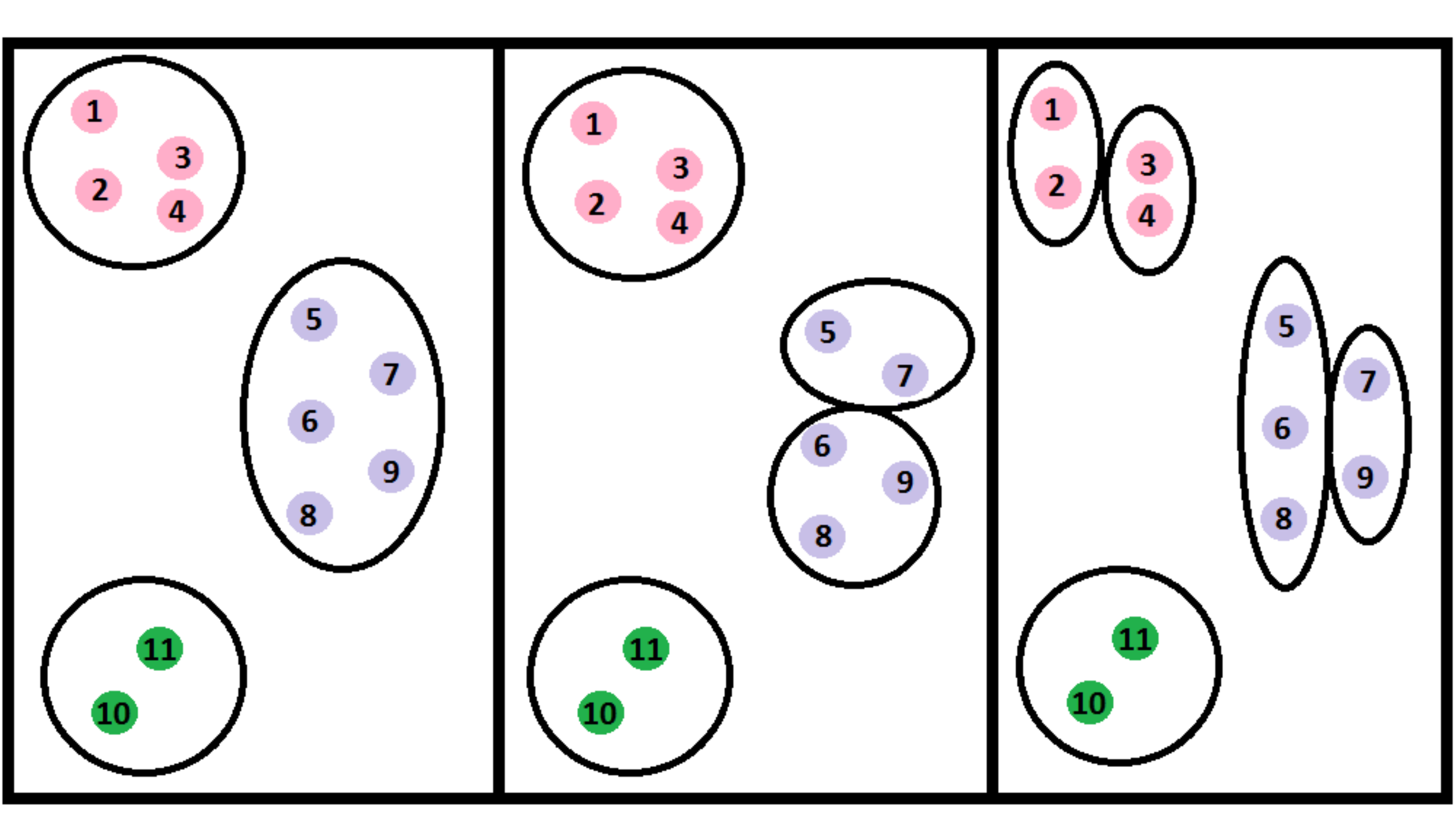}
	\caption{Clusters}
	\label{fig:cluster}
        \end{subfigure}
        \begin{subfigure}[b]{0.43\textwidth}
	\includegraphics[width=\textwidth]{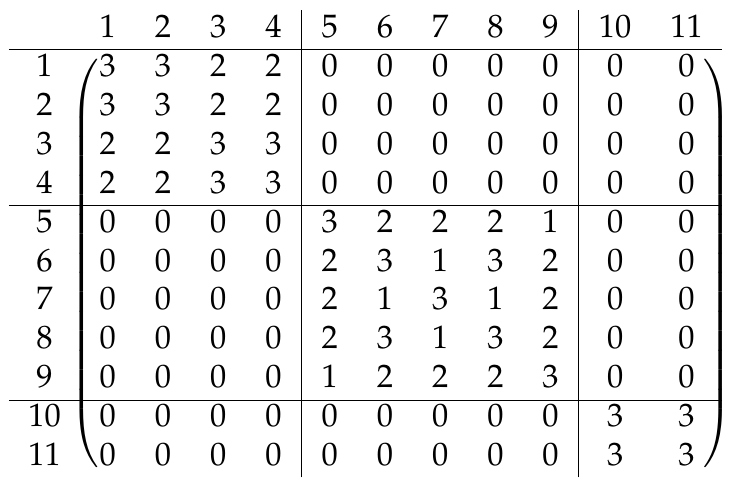}
	\caption{Consensus Matrix C}
	\label{fig:matrix}
        \end{subfigure}
        \caption{Consensus Clustering}
        \vspace{-10pt}
\end{figure}

Consensus clustering combines the advantages of many algorithms to find a better clustering. 
Different algorithms are run on the same dataset and a consensus matrix is created such that each time data points $i$ and $j$ are clustered together, a 1 is added to the consensus matrix at positions $ij$ and $ji$.
It should be noted that a consensus matrix can be created by running the same algorithm, such as $k$-means multiple times with varying parameters, such as number of clusters.
In the case of text mining, the consensus matrix is then used in place of the term document matrix when clustering again. 
Figure ~\ref{fig:cluster} shows the results of three different clustering algorithms.
Note that data points $1$ and $3$ cluster together two out of three times  \cite{Race}.
Thus in  Figure ~\ref{fig:matrix} there is a $2$ at position $C_{1,3}$ and $C_{3,1}$.

\FloatBarrier

\vspace{0.2in}

\subsection{Non-Negative Matrix Factorization}

 Non-Negative Matrix Factorization (NMF) decomposes the term-document matrix into two matrices: a term-topic matrix and a topic-document matrix with $k$ topics. 
 The term document matrix $A$ is decomposed such that

\[
A \approx WH
\] where $A$ is an $m$ x $n$ matrix, $W$ is an $m$ x $k$ non-negative matrix, and $H$ is a $k$ x $n$ non-negative matrix.

\vspace{0.2in}

Each topic vector is a linear combination of words in the text dictionary.
Each document or column in the term document matrix can be written as a linear combination of these topic vectors such that

\[
A_j = h_{1j}w_1 + h_{2j}w_2 + \cdots + h_{kj}w_k
\] where $h_{ij}$ is the amount that document $j$ is pointing in the direction of topic vector $w_i$ \cite{NMFClus}  \nocite{NMFemail} .

\vspace{0.2in}

The Multiplicative Update Rule, Alternating Least Squares (ALS), and Alternating Constrained Least Squares (ACLS) algorithms are three of the most widely used algorithms that calculate $W$ and $H$ and aim to minimize $\lVert A - WH \rVert$. 

\vspace{0.2in}

The biggest advantage of the Multiplicative Update Rule is that, in theory, it converges to a local minimum. However, the initialization of $W$ and $H$ can greatly influence this minimum \cite{LeeSeung}. One problem with the Multiplicative Update Rule is that it can be time costly depending on how $W$ and $H$ are initialized. Further, the two matrices have no sparsity, and the 0 elements in the $W$ and $H$ matrices are locked, meaning if an element in $W$ or $H$ becomes 0, it can no longer change. In text mining, this results in words being removed from but not added to topic vectors. Thus, once the algorithm starts down a path for the local minimum, it cannot easily change to a different one even if that path leads to a poor topic vector \cite{MeyerLangvilleNMF}.   

\begin{algorithm}
\caption{Multiplicative Update Rule}
\label{alg:mult}
\begin{algorithmic}[1]

\STATE \textbf{Input:} $A$ term document matrix ($m$ x $n$), $k$ number of topics

\STATE $W$ = abs(rand($m$,$k$))

\STATE $H$ = abs(rand($k$,$n$))

\FOR{$i$ = 1:maxiter}

\STATE $H = H .* (W^TA) ./ (W^TWH +10^{-9})$

\STATE $W = W .* (AH^T) ./ (WHH^T+10^{-9})$

\ENDFOR
\end{algorithmic}
\end{algorithm}

One of the biggest advantages of the Alternating Least Square algorithm is its speed of convergence. Another is that only matrix $W$ is initialized and matrix $H$ is calculated from $W$'s initialization. The 0 elements in matrices $W$ and $H$ are not locked; thus, this algorithm is more flexible in creating topic vectors than the Multiplicative Update Rule. The biggest disadvantage of this algorithm is its lack of sparsity in the $W$ and $H$ matrices \cite{MeyerLangvilleNMF}. 

\begin{algorithm}
\caption{Alternating Least Square}
\label{alg:als}
\begin{algorithmic}[1]

\STATE \textbf{Input:} $A$ term document matrix ($m$ x $n$), $k$ number of topics

\STATE $W$ = abs(rand($m$,$k$))

\FOR{$i$ = 1:maxiter}

\STATE solve $W^TWH = W^TA$ for $H$

\STATE replace all negative elements in $H$ with 0

\STATE solve $HH^TW^T = HA^T$ for $W$

\STATE replace all negative elements in $W$ with 0

\ENDFOR
\end{algorithmic}
\end{algorithm}

The Alternating Constrained Least Square algorithm has the same advantages as the Alternating Least Square algorithm with the added benefit that matrices $W$ and $H$ are sparse. This algorithm is the fastest of the three, and since our data is fairly large, we used the ACLS algorithm every time we performed NMF on our Twitter data \cite{MeyerLangvilleNMF}.

\begin{algorithm}
\caption{Alternating Constrained Least Square}
\label{alg:acls}
\begin{algorithmic}[1]

\STATE \textbf{Input:} $A$ term document matrix ($m$ x $n$), $k$ number of topics

\STATE $W$ = abs(rand($m$,$k$))

\FOR{$i$ = 1:maxiter}

\STATE solve $(W^TW + \lambda_H I)H = W^TA$ for $H$

\STATE replace all negative elements in $H$ with 0

\STATE solve $(HH^T + \lambda_W I)W^T = HA^T$ for $W$

\STATE replace all negative elements in $W$ with 0

\ENDFOR
\end{algorithmic}
\end{algorithm}

\subsection{DBSCAN}

Density-Based Spatial Clustering of Applications with Noise (DBSCAN) is a common clustering algorithm. 
DBSCAN uses some similarity metric, usually in the form of a distance, to group data points together.
DBSCAN also marks points as noise, so it can be used in noise removal applications.

\begin{figure}[h]
        \centering
	 \includegraphics[scale=0.5]{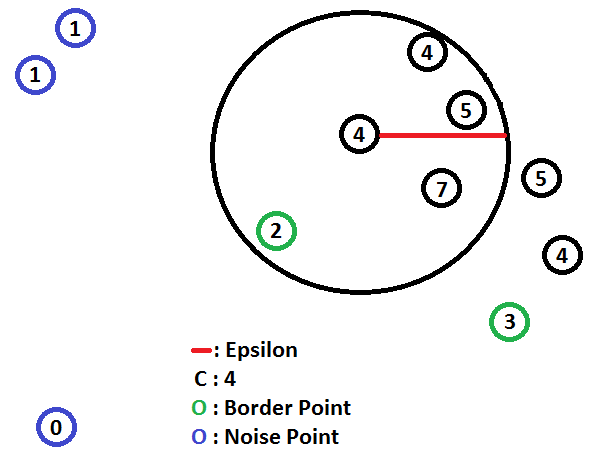}
	\caption{Noise Points in DBSCAN}
        \vspace{-10pt}
\end{figure}
\FloatBarrier

DBSCAN requires two inputs:  $c$ minimum number of points in a dense cluster, and $\epsilon$ distance. 
DBSCAN visits every data point in the dataset and draws an $\epsilon$ radius around the point. 
If there is at least $c$ number of points in the $\epsilon$ radius, we call the point a dense point. 
If there are not the $c$ minimum number of points in the $\epsilon$ radius, but there is a dense point, then we call the point a border point. 
Finally, if there is neither the $c$ number of points nor a dense point in the radius, we call the point a noise point. 
In this way, DBSCAN can be used to remove noise. \cite{Ester96adensity-based} 

\vspace{0.2in}

DBSCAN has a few weaknesses. 
First, it is highly dependent on its parameters. 
Changing $\epsilon$ or $c$ will drastically change the results of the algorithm. 
Also, it is not very good at finding clusters of varying densities because it does not allow for a variation in $\epsilon$.

\vspace{0.2in}

\section{Methods}

\subsection{Removing Retweets}

The data we analyzed were tweets from Twitter containing the words `world cup'.
Many of the tweets were the same; they were what Twitter calls  a `retweet'.
Since a retweet does not take as much thought as an original tweet we decided to remove the retweets as to prevent a bias in the data.
If several columns in our term document matrix were identical, we removed all but one of those columns. 
This process removed about 9,000 tweets from the data.
In our preliminary exploration of the tweets, we found a topic about the Harry Potter Quiddich World Cup. 
When we looked at the tweets that were contained in this cluster, we found that they were all the same tweet, retweeted approximately 2,000 times. 
When these retweets were removed, the cluster no longer existed because it was reduced to 1 tweet about Quiddich. 
Thus, we found that removing retweets eliminated clusters that only contained a small number of original tweets.

\subsection{Removing Noise in World Cup Tweets}

\begin{figure}[h]
        \centering
        \begin{subfigure}[b]{0.48\textwidth}
	\includegraphics[width=\textwidth]{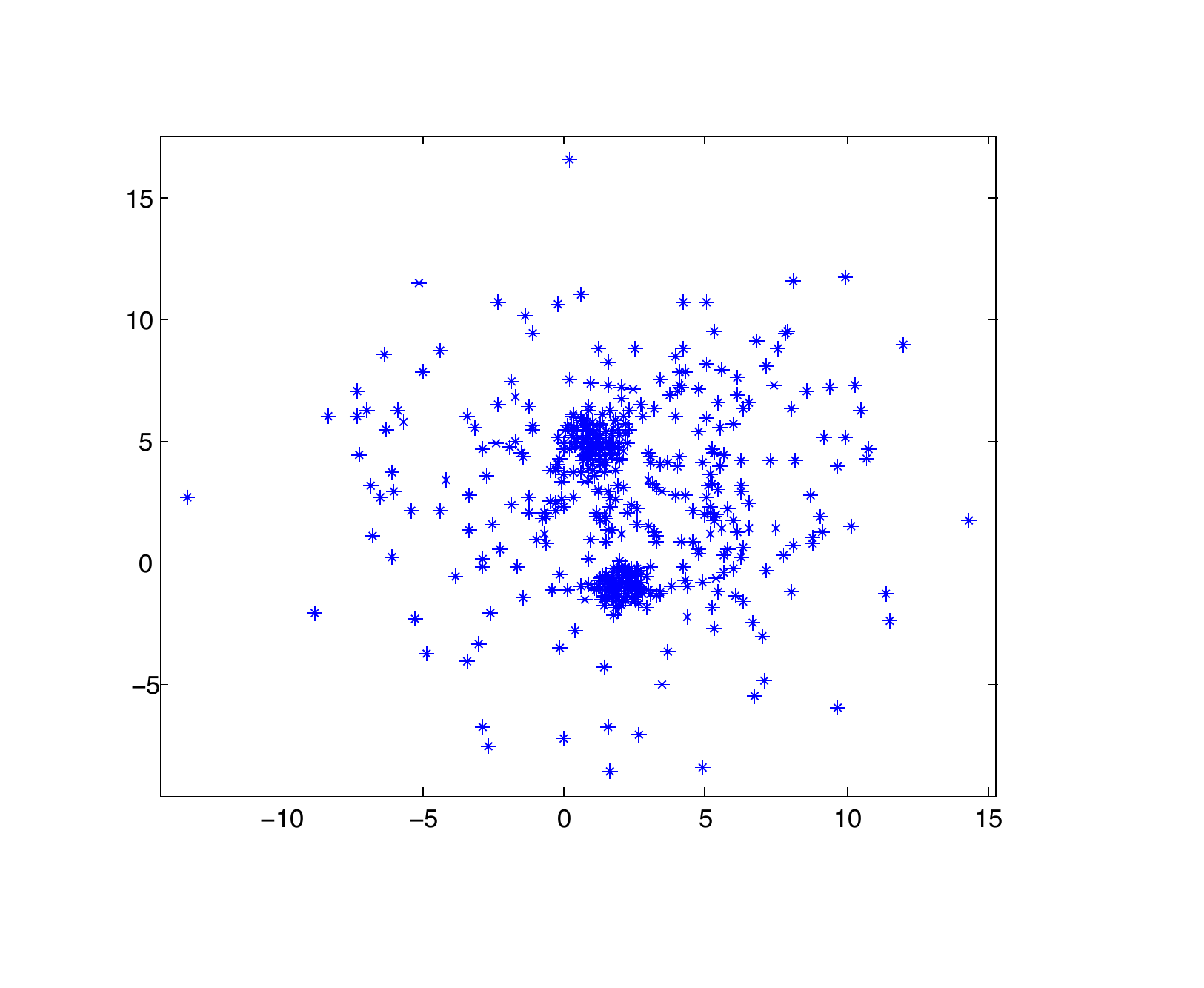}
	\caption{Before Noise Removal}
        \end{subfigure}
        \begin{subfigure}[b]{0.48\textwidth}
	\includegraphics[width=\textwidth]{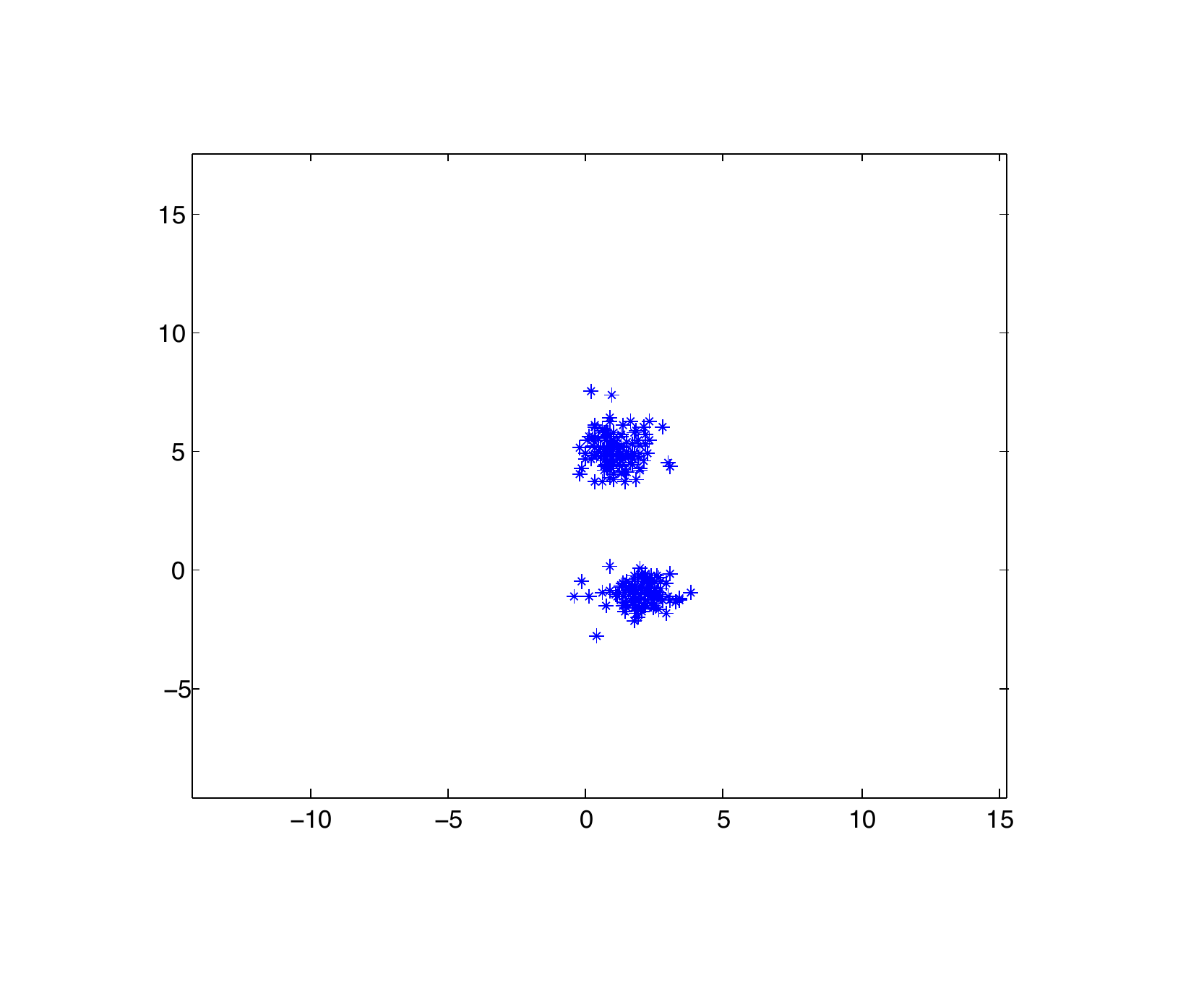}
	\caption{After Noise Removal}
        \end{subfigure}
        \caption{Noise Removal}
        \label{fig:NoiseRemove}
        \vspace{-10pt}
\end{figure}

Tweets are written and posted without much revision.
That is to say that tweets will contain noise.
Some of that noise in the vocabulary of tweets can be removed with a stop list and by stemming.
When we look at a collection of tweets we want the tweets that are the most closely related to one specific topic.
Tweets on the edges of clusters are still related to the topic just not as closely.
Therefore, we can remove them as noise without damaging the meaning of the cluster.
Figure ~\ref{fig:NoiseRemove} shows a simple two-dimensional example of how the noise is removed while still keeping the clusters.
We created four algorithms for noise removal.

\FloatBarrier

The first algorithm that we created used only the consensus matrix created by multiple runs of $k$-means where the $k$ value varied.
We wanted to vary $k$ so that we could see which tweets clustered together more frequently.
These were then considered the clusters, and other points were removed as noise.
We did this by creating a drop tolerance on the consensus matrix.
If tweets $i$ and $j$ did not cluster together more than $10\%$ of the time, term $ij$ in the consensus matrix was dropped to a 0.
Then we looked at the row sums for the consensus matrix and employed another drop tolerance. 
All the entries in the consensus matrix were averaged.
Tweets whose row sum was less than that average were marked as noise points.
The problem with this algorithm is that the clusters must be of similar density.
When there is a variation in the density of clusters the less dense cluster is removed as noise.

\vspace{0.2in}

The second algorithm used multiple runs of DBSCAN that helped us decide if a tweet was a true noise point or not.
The distance matrix that we used was based on the cosine distance between tweets.
We used the cosine distance because it is standard when looking at the distance between text data.
Since our first algorithm removed less dense clusters, we wanted to make sure that they were still included and not removed as noise.
As DBSCAN is so dependent on the $\epsilon$, we used a range of $\epsilon$ in order to include those clusters.
Through experimentation we found that larger data sets required more runs of DBSCAN.
We created a matrix that was the number of tweets by the number of runs of DBSCAN where each entry $ij$ in the matrix was the classification: dense, border, or noise point, for the $i^{th}$ tweet on the $j^{th}$ run.
If the tweet was marked as a border point or noise point by more than $50\%$ of the runs it was considered a true noise point.
We also looked into varying $c$.
However, this created problems as the algorithm then marked all the tweets as noise points.
Therefore, we decided to keep the $c$ value constant.
While this algorithm kept clusters of varying density it was more difficult to tell the clusters apart.

\vspace{0.2in}

Since the consensus matrix is a similarity matrix, we decided to use DBSCAN on that matrix instead of a distance matrix.
The idea is similar to the second algorithm; we still varied $\epsilon$ and kept $c$ constant.
The $\epsilon$ value in this algorithm was now the number of times tweet $i$ and $j$ clustered together.
We performed DBSCAN multiple times on the consensus matrix and created a new matrix of classification as described before.
Again we decided that if the tweet was marked as a border point or noise point by more than $50\%$ of runs it was considered a true noise point.
This algorithm is unique because it removes noise points between the clusters.
We found that this is because the points between clusters will vary more frequently in which cluster they belong.

\vspace{0.2in}

We wanted to use all the strengths from the previous algorithms so we combined them.
This new algorithm looks at the classification from each of the previous algorithms where a noise point is represented by a $0$.
Then if at least two of the three algorithms marked a tweet as a noise point it would be removed from the data.
This allows us to remove points on the edge and between clusters but still keep clusters of varying density. 
This process removed about 3,558 tweets from the data.
This was the final dataset that we used in order to find major topics in the tweets. 

\subsection{Choosing a number of Topics}

\begin{figure}[h]
\centering
\includegraphics[scale=0.5]{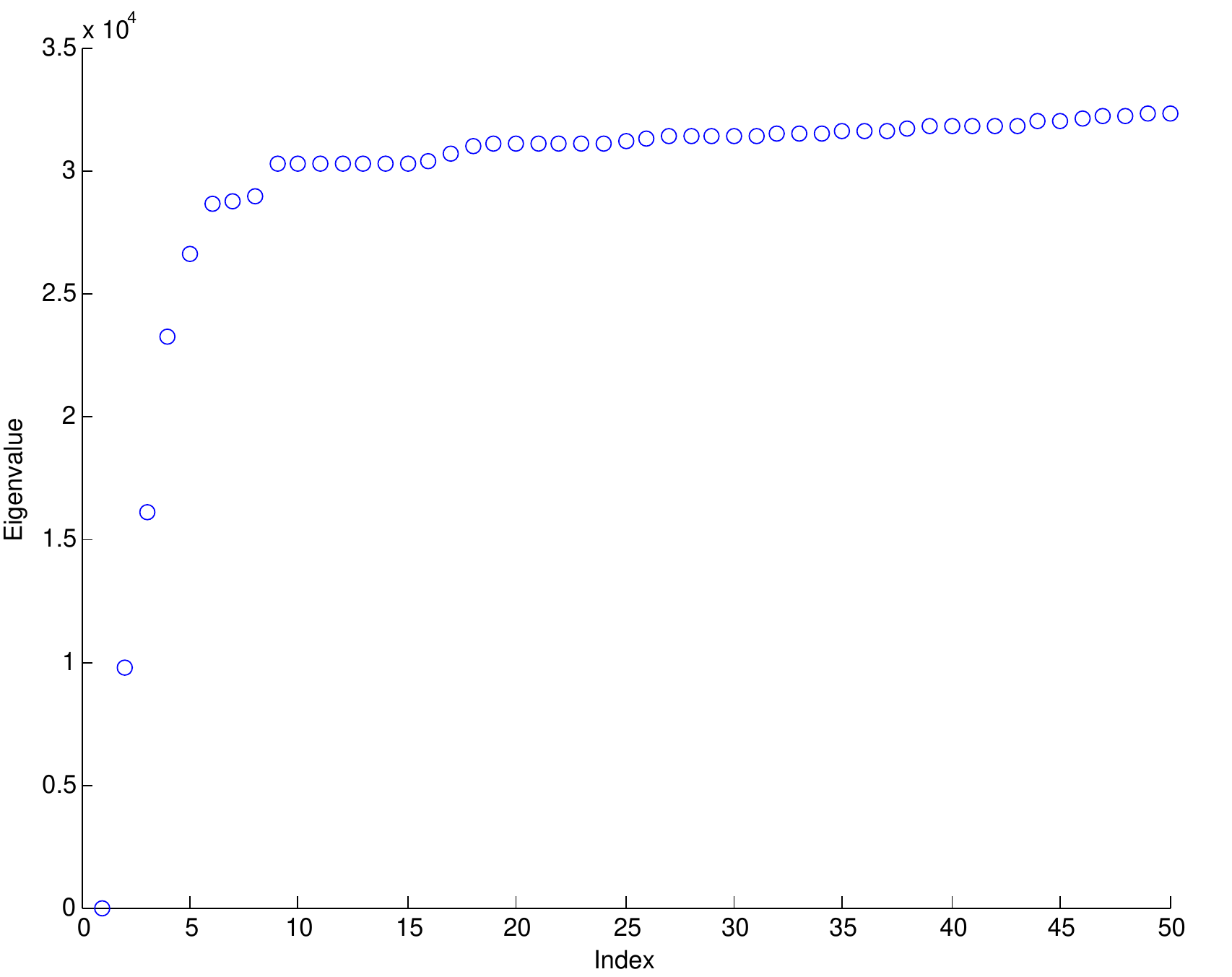}
\caption{Eigenvalues of Laplacian Matrix}
\label{fig:Evalues}
\end{figure}

To decide how many topics we would ask the algorithms to find, we created the Laplacian matrix ($L$) such that $$L = D - C$$ where $D$ is a diagonal matrix with entries corresponding to the sum of the rows of the consensus matrix, $C$ \cite{NewmanModularity}.
We looked at the 50 smallest eigenvalues of the Laplacian matrix to identify the number of topics we should look for.
A gap in the eigenvalues signifies the number of topics.
There are large gaps between the first 6 eigenvalues, but we thought that a small number of topics would make the topics too broad.
Since we wanted a larger number of topics we chose to use the upper end of the gap between the $8^{th}$ and $9^{th}$ eigenvalues as shown in Figure ~\ref{fig:Evalues}.
For future work, we would like to use a normalized Laplacian matrix to create a better eigenvalue plot. 

\FloatBarrier

\subsection{Clustering World Cup Tweets with Consensus Matrix}
 
We clustered the remaining tweets in order to find major themes in the text data. 
Since $k$-means is the most widely used clustering algorithm and since its results are highly dependent on the value of $k$, we ran $k$-means on our Twitter data with $k = 2$ through $12$. 
We then ran $k$-means a final time with the consensus matrix as our input and $k = 9$. 
The algorithm gave us the cluster to which each tweet belonged and we placed each tweet in a text file with all the other tweets from its cluster.
 Then we created a word cloud for each cluster in order to visualize the overall themes throughout the tweets. 

\subsection{Clustering World Cup Tweets with Non-Negative Matrix Factorization}

The problem with using the $k$-means algorithm is that the only output from $k$-means is the cluster number of each tweet.
 Knowing which cluster each tweet belonged to did not help us know what each cluster was about. 
 Although it is possible to look at each tweet in each cluster and determine the overall theme, it usually requires some visualization tools, such as a word cloud, in order to discover the word or words that form a cluster. 
 Thus, we used a Non-Negative Matrix Factorization algorithm, specifically the Alternating Constrained Least Square (ACLS) algorithm, in order to more easily detect the major themes in our text data.

The algorithm returns a $W$ term-topic matrix and an $H$ topic-document matrix. 
We ran ACLS with $k = 9$ and sorted the rows in descending order such that the first element in column $j$ corresponded with the most important word to topic $j$. 
Thus, it was possible to see the top 10 or 20 most important words for each of the topics.  

Once we found the most important words for each topic, we were curious to see how these words fit together. 
We created an algorithm that picked a representative tweet for each topic such that the representative tweet had as many words from the topic as possible. 
We called these tweets topic sentences.

\section{Results}

\begin{figure}[h]
        \centering
        \includegraphics[scale=0.5]{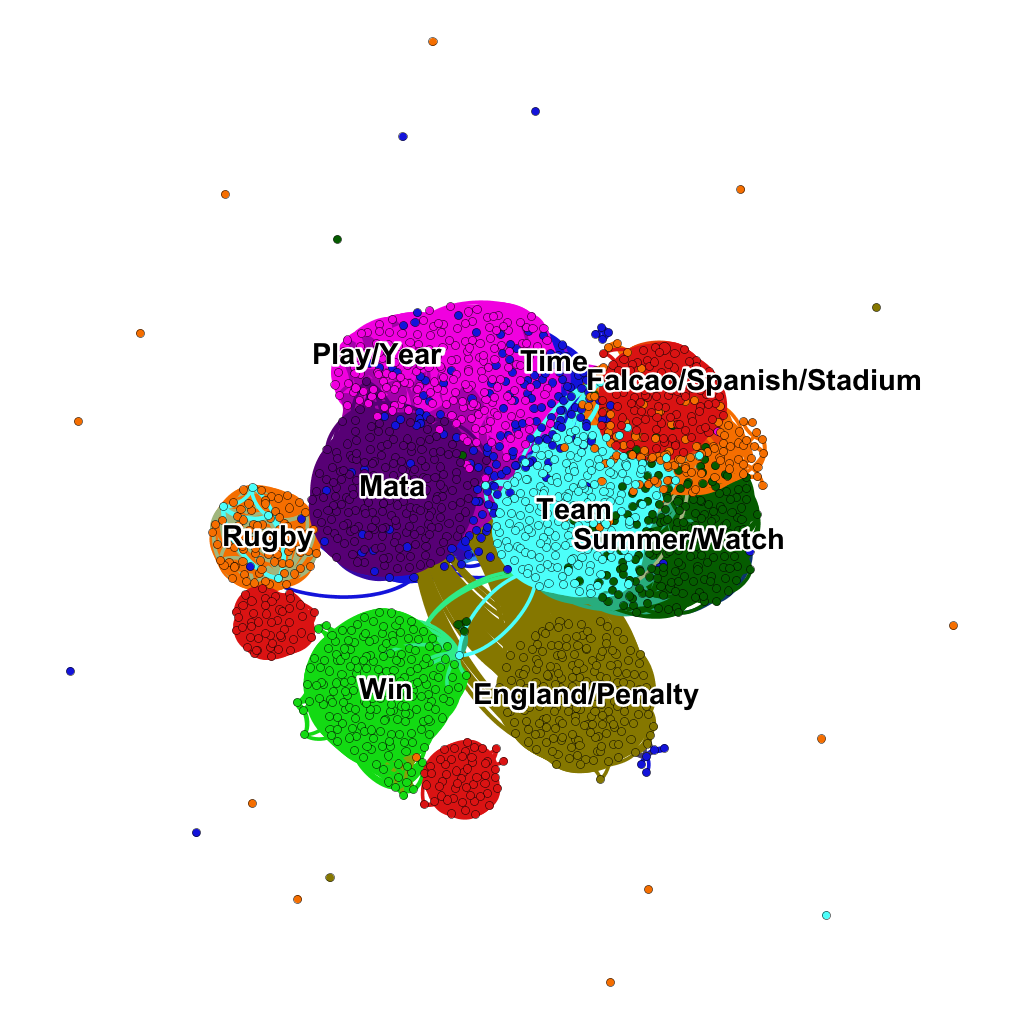}
        \caption{Topics found by $k$-means}
        \label{fig:KmeansG}
        \vspace{-0.18in}
\end{figure}

In the visualizations from the graphing software Gephi we are able to see how close topics are to one another.
In the graph of the consensus matrix, two tweets are connected if they are clustered together more than 8 times.
If the tweets are clustered together more frequently, they are closer together in the graph and form a topic, represented by a color in the graph.
The unconnected nodes are tweets that are not clustered with any other tweet more than 8 times.
Because of the way $k$-means works we see that some of the topics are split in Figure \ref{fig:KmeansG}.
The most obvious split is the `Falcao/Spanish/Stadium' topic.

\FloatBarrier

\begin{figure}[h]
        \centering
        \includegraphics[scale=0.45]{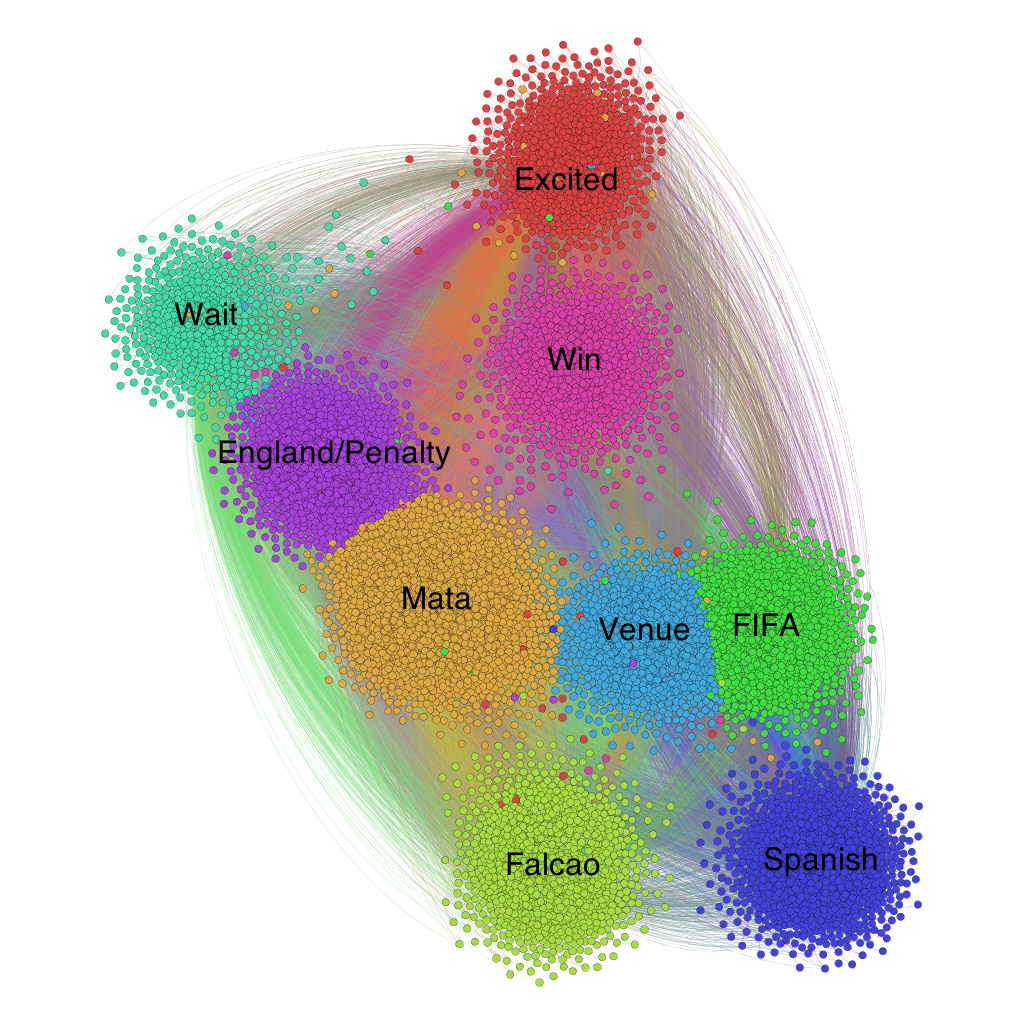}
        \caption{Topics found by NMF}
        \label{fig:NMFG}
\end{figure}

In the graph for NMF, the colored nodes represent the topic that the tweet is most closely related to.
The edges emanating from a node represent the other topics that the tweet is only slightly related to.
We created the graph in such a way that distance between heavier weights is shorter.
This pulls topics that are similar towards each other.
For example, the `FIFA' and `Venue' topics are right next to each other as seen in Figure ~\ref{fig:NMFG}.
This means that there are tweets in the `FIFA' topic that are highly related to the `Venue' topic.
When we further examined these two topics, we found that they both shared the words `stadium' and `Brazil' frequently.

\FloatBarrier

We wanted to compare the results from $k$-means and NMF so we created visualizations of the most frequent words with software called Wordle.
For example, NMF selected the Spanish tweets as their own topic.
When we looked for the same topic in the results of $k$-means we found that it created one cluster that contained the Spanish topic, a topic about the player Falcao, and a topic about stadiums. 
From this we thought that NMF was more apt at producing well defined clusters.

\begin{figure}[h]
        \centering
        \includegraphics[scale=0.5]{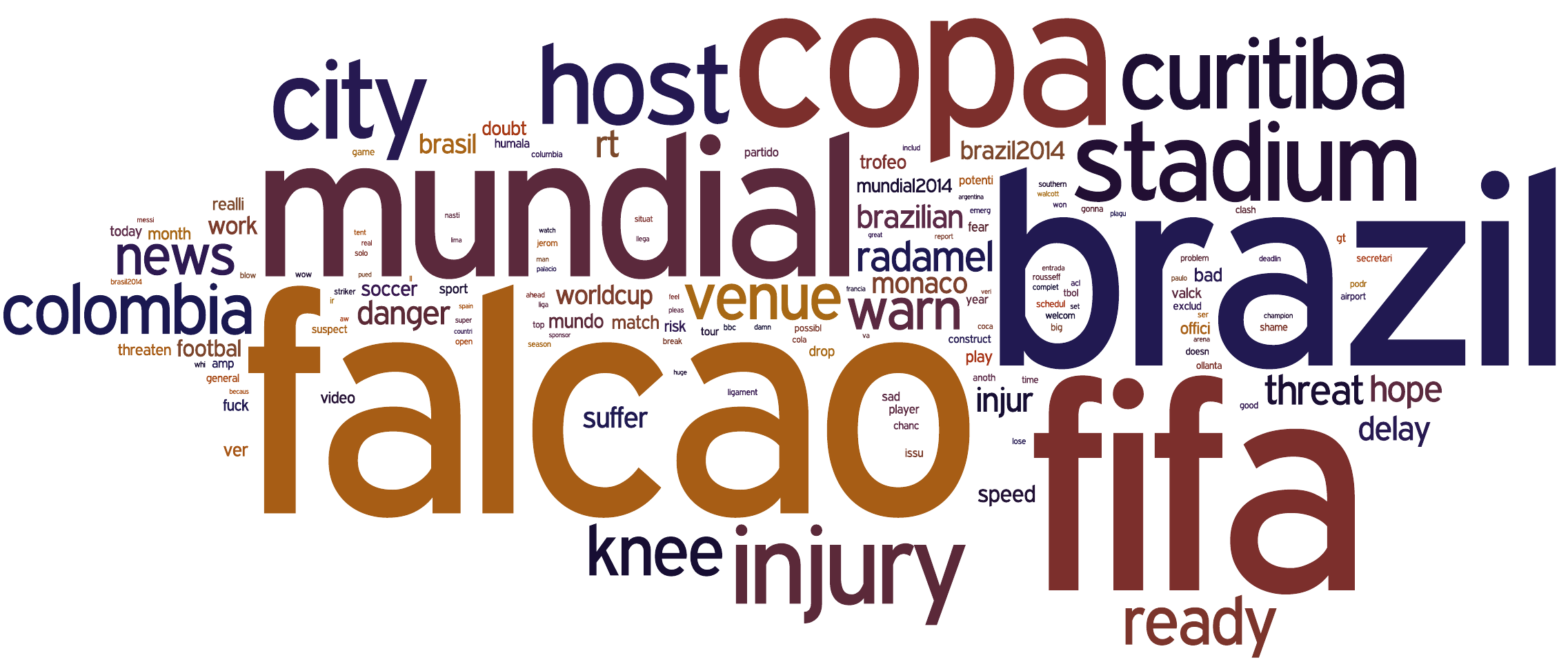}
        \caption{Falcao/Spanish/Stadium Topic from $k$-means}
        
        \begin{subfigure}[b]{0.47\textwidth}
	\includegraphics[width=\textwidth]{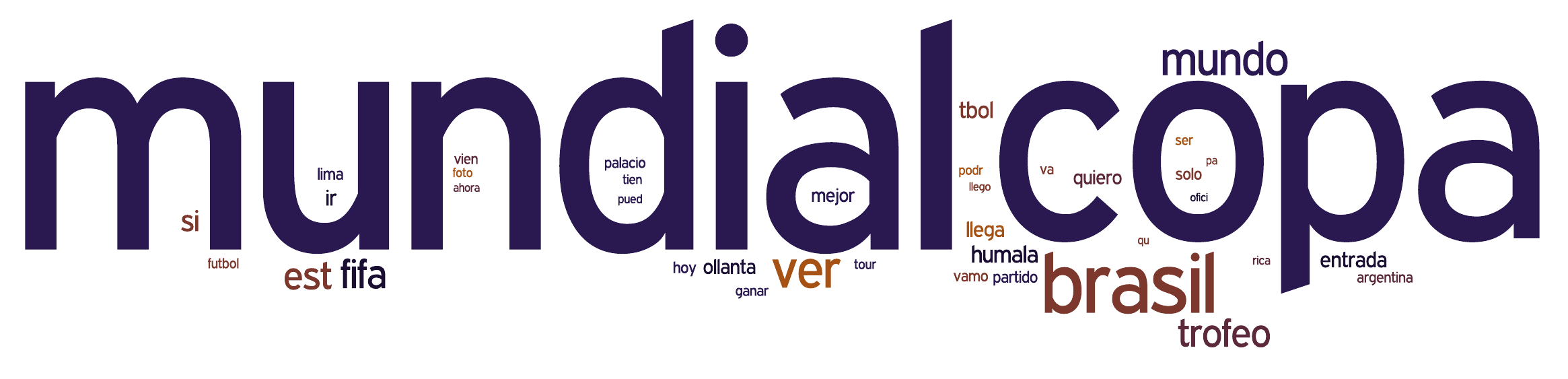}
	\caption{Spanish Topic from NMF}
        \end{subfigure}
        \quad
        \begin{subfigure}[b]{0.47\textwidth}
	\includegraphics[width=\textwidth]{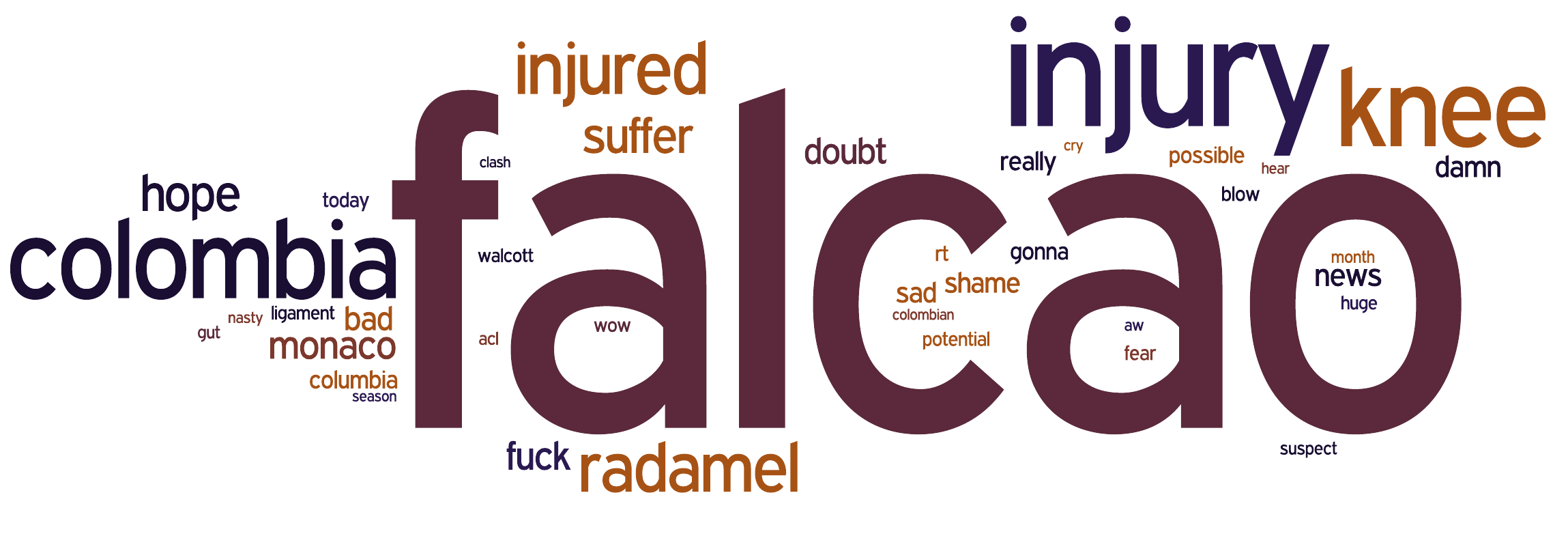}
	\caption{Falcao Topic from NMF}
        \end{subfigure}
        
        \begin{subfigure}[b]{0.47\textwidth}
	\includegraphics[width=\textwidth]{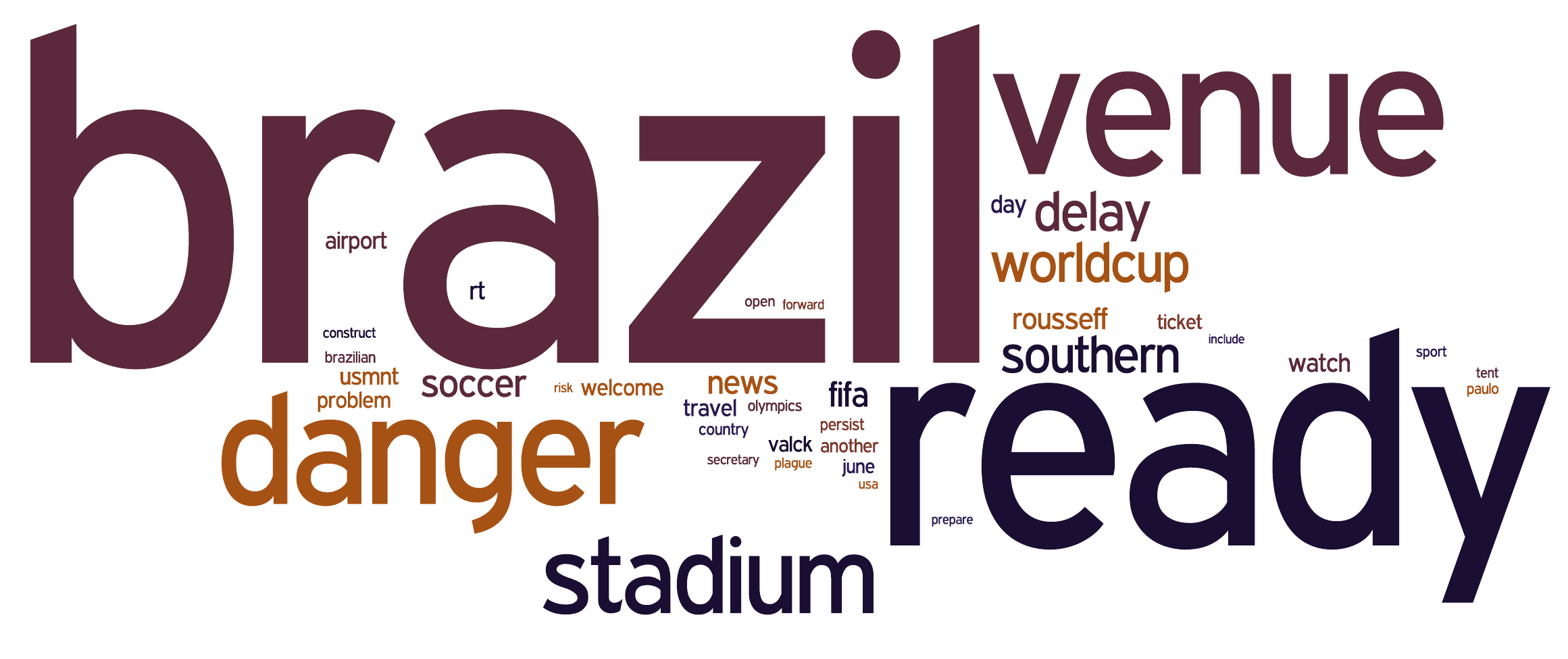}
	\caption{Venue Topic from NMF}
        \end{subfigure}
        \quad
        \begin{subfigure}[b]{0.47\textwidth}
	\includegraphics[width=\textwidth]{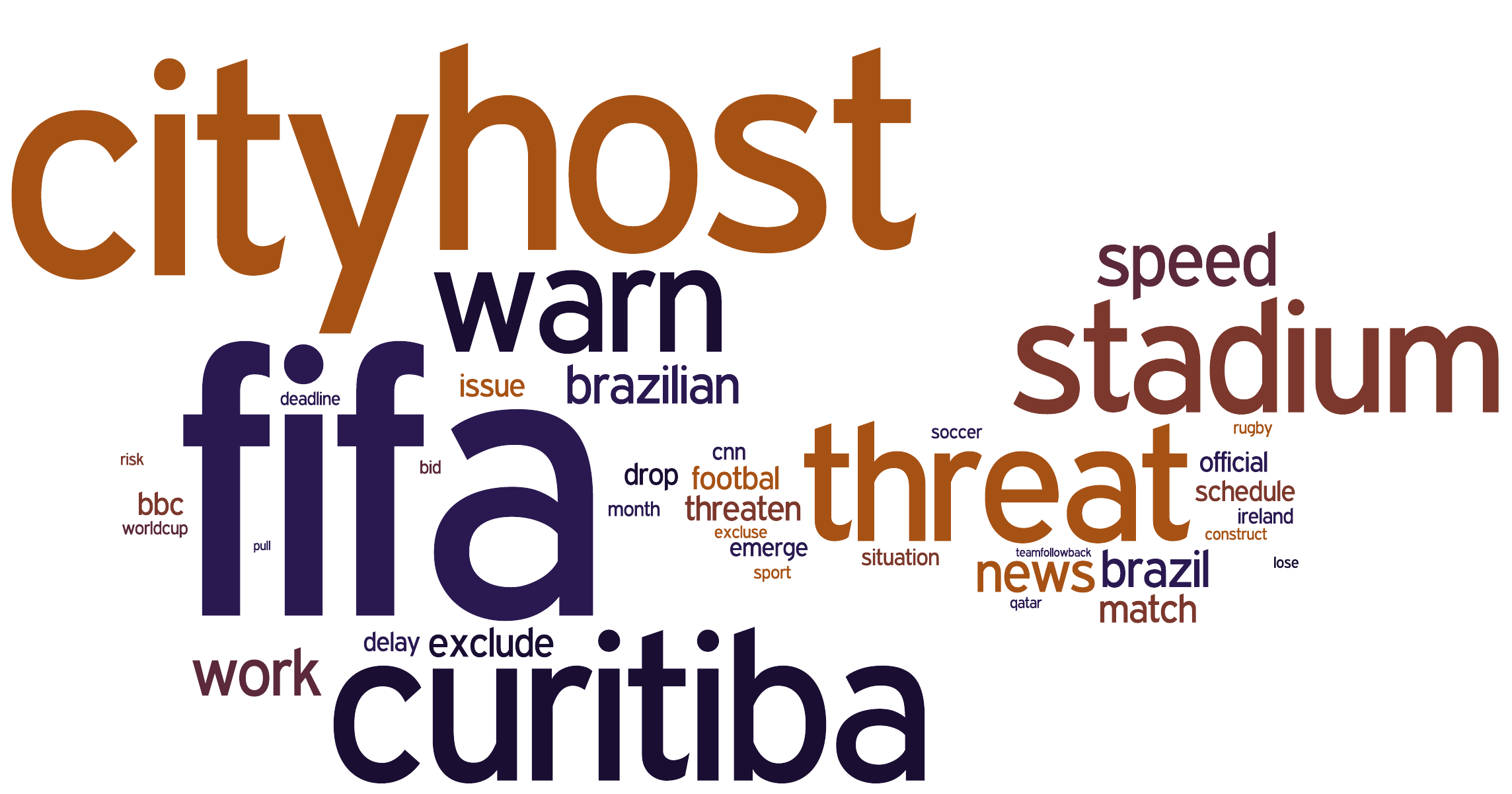}
	\caption{FIFA Topic from NMF}
        \end{subfigure}
        \caption{NMF topics that create $k$-means topic}
 \vspace{-0.2in}
\end{figure}
\FloatBarrier

We conclude that NFM is a better algorithm for clustering these tweets.
NMF was computationally faster and provided more specific topics than $k$-means.

\section{Conclusion} 
We used cluster analysis to find topics in the collection of tweets.
NMF proved to be faster and provided more easily interpreted results.
NMF selected a single tweet that represented an entire topic whereas $k$-means can only provide the tweets in each topic.
Further visualization techniques are necessary for interpreting the meanings of the clusters provided by $k$-means. 

\vspace{0.2in}

There is still more to explore with understanding text data in this manner.
We only looked at NMF and $k$-means to analyze these tweets.
Other algorithms that we did not use could prove to be more valuable.
Since we only looked deeply into text data, further research could prove that other algorithms are better for different types of data.
We explored our results using two visualization tools, Gephi and Wordle. 
There is still much to be done in this aspect.
In retrospect we would perform Singular Value Decomposition \cite{MeyerTB} on our consensus matrix before running $k$-means.
This way noise would be removed and the clustering would be more reliable.
For those interested in further exploration along the lines of our case study, a natural extension would be to perform the analysis in real-time so as to observe how specific topics evolve with time.

\section{Acknowledgments}
We would like to thank the National Science Foundation and National Security Agency for funding this project. 
We express gratitude to the NC State Research Experience for Undergraduates in Mathematics: Modeling and Industrial Applied Mathematics program for providing us with this opportunity.



\bibliographystyle{plain}
\bibliography{bibliography3}

\end{document}